%% file: acl_latex.tex
\pdfoutput=1

\documentclass[11pt]{article}

\usepackage{acl}

\usepackage{times}
\usepackage{latexsym}

\usepackage[T1]{fontenc}

\usepackage[utf8]{inputenc}

\usepackage{microtype}

\usepackage{inconsolata}

\usepackage[ruled,noline,nofillcomment]{algorithm2e}
\usepackage{amsfonts}
\usepackage{amsmath}
\usepackage{bbding}
\usepackage{graphicx}
\usepackage{longtable}
\usepackage{makecell}
\usepackage{arydshln}
\usepackage{multirow}
\usepackage{ulem}

\title{CoE-SQL: In-Context Learning for Multi-Turn Text-to-SQL with Chain-of-Editions}

\author{Hanchong Zhang$^{1}$, Ruisheng Cao$^{1}$, Hongshen Xu$^{1}$, Lu Chen$^{1,2*}$ and Kai Yu$^{1,2}$\thanks{\ \ The corresponding authors are Lu Chen and Kai Yu.} \\
$^{1}$X-LANCE Lab, Department of Computer Science and Engineering \\
MoE Key Lab of Artificial Intelligence, SJTU AI Institute \\
Shanghai Jiao Tong University, Shanghai, China \\
$^{2}$Suzhou Laboratory, Suzhou, China \\
{\tt \{zhanghanchong,chenlusz,kai.yu\}@sjtu.edu.cn}}

\begin{document}
\maketitle
\begin{abstract}
Recently, Large Language Models (LLMs) have been demonstrated to possess impressive capabilities in a variety of domains and tasks. We investigate the issue of prompt design in the multi-turn text-to-SQL task and attempt to enhance the LLMs' reasoning capacity when generating SQL queries. In the conversational context, the current SQL query can be modified from the preceding SQL query with only a few operations due to the context dependency. We introduce our method called CoE-SQL\footnote{Our code is publicly available at \url{https://github.com/X-LANCE/text2sql-multiturn-GPT}} which can prompt LLMs to generate the SQL query based on the previously generated SQL query with an edition chain. We also conduct extensive ablation studies to determine the optimal configuration of our approach.
Our approach outperforms different in-context learning baselines stably and achieves state-of-the-art performances on two benchmarks SParC and CoSQL using LLMs, which is also competitive to the SOTA fine-tuned models.
\end{abstract}

\input{contents/introduction}

\input{contents/related_work}

\input{contents/methodology}

\input{contents/experiments}

\input{contents/conclusion}

\input{contents/limitations}

\input{contents/acknowledgements}

\normalem
\bibliography{anthology,custom}

\appendix
\onecolumn

\input{contents/appendix_a}

\input{contents/appendix_b}

\input{contents/appendix_c}

\input{contents/appendix_d}

\end{document}

%% file: contents/introduction.tex
\section{Introduction}
Text-to-SQL~\cite{zhongSeq2SQL2017, xu2017sqlnet} is a semantic parsing task that translates the natural language question into the appropriate SQL query according to the given database schema. This technique is critical in building a natural language interface to relational databases~\cite{nlidb}, which alleviates the burden on programmers to curate valid and correct annotations.

In this work, we focus on the contextual settings~\cite{yu-etal-2019-sparc,yu-etal-2019-cosql} where users interact with the system in multi-turn scenarios. In each turn, the text-to-SQL parser understands and interprets the current user question into a SQL query based on the entire dialogue history. Considering the requirement of context modeling, EditSQL~\cite{zhang-etal-2019-editing} introduces copy mechanism during the generation of SQL queries to re-use the SQL spans in history. DELTA~\cite{delta} firstly paraphrases the long context into a single question and transforms the original problem into single-turn parsing. IGSQL~\cite{cai-wan-2020-igsql} and HIESQL~\cite{zheng-etal-2022-hie} both utilize the concept of cross-turn schema interaction into the specialized design of graph encoders. STAR~\cite{star} and CQR-SQL~\cite{cqrsql} inject the capability of context modeling through self-supervised pre-training tasks. Though remarkable progress has been achieved, these task-specific methods rely on complicated model design which is labor-intensive and training-inefficient.

The advent of LLMs~\cite{ouyang2022training} provides a promising solution in the text-to-SQL field, that is in-context learning with few-shot exemplars~\cite{brown2020language}. In this branch, DIN-SQL~\cite{pourreza2023dinsql} adopts the concept of Chain-of-Thought~(CoT, \citealp{wei2023chainofthought}) and establishes a new milestone in single-turn settings. ACT-SQL~\cite{zhang2023actsql} further eliminates the necessity of mannually crafting CoT exemplars and automatically constructs useful demonstraions. However, adapting the ICL method into contextual settings is less investigated. ACT-SQL attempts to reduce the original problem into single-turn problems via sentence rewriting. But poor performances are reported on multi-turn benchmarks due to severe error propagation.

\begin{table}[htbp]
\centering
\resizebox{0.49\textwidth}{!}{
\begin{tabular}{p{\columnwidth}}
\hline

\hline
\textbf{Turn 1} \\
\textbf{Q:} {\it What are the phone names available in Alberta?} \\
\textbf{SQL:} {\small SELECT T1.Name FROM phone AS T1 JOIN phone\_market AS T2 JOIN market AS T3 ON T1.Phone\_ID = T2.Phone\_ID AND T2.Market\_ID = T3.Market\_ID WHERE T3.District = "Alberta"} \\
\textbf{Edit:} no edition \\
\hdashline
\textbf{Turn 2} \\
\textbf{Q:} {\it Show the district as well.} \\
\textbf{SQL:} {\small SELECT T1.Name, \textcolor[RGB]{0,112,192}{\underline{T3.District}} FROM phone AS T1 JOIN phone\_market AS T2 JOIN market AS T3 ON T1.Phone\_ID = T2.Phone\_ID AND T2.Market\_ID = T3.Market\_ID WHERE T3.District = "Alberta"} \\
\textbf{Edit:} add select column \textcolor[RGB]{0,112,192}{T3.District}\\
\hdashline
\textbf{Turn 3} \\
\textbf{Q:} {\it Remove the restriction of those in Alberta only, show all instead.} \\
\textbf{SQL:} {\small SELECT T1.Name, T3.District FROM phone AS T1 JOIN phone\_market AS T2 JOIN market AS T3 ON T1.Phone\_ID = T2.Phone\_ID AND T2.Market\_ID = T3.Market\_ID \textcolor[RGB]{176,23,31}{\sout{WHERE T3.District = "Alberta"}}} \\
\textbf{Edit:} delete where clause \textcolor[RGB]{176,23,31}{T3.District = "Alberta"}\\
\hline

\hline
\end{tabular}
}
\caption{A multi-turn example from SParC~\cite{yu-etal-2019-sparc}. Each edition is based on the previous turn.}
\label{tab:edit_case}
\end{table}

In this work, we propose Chain-of-Edition framework to tackle the multi-turn text-to-SQL task~(CoE-SQL). Inspired by recent progress in code edition~\cite{chen-etal-2023-sqledit}, we attempt to model the changes in the output SQL program across different interaction turns explicitly. These changes can be serialized into the reasoning thought, as a specific implementation of CoT, to track the shift in the user focus and intention. 
The underlying mechanism is that, in a coherent and consistent dialogue, the user's questions often depend on the previous focus, and the latest request or intent can be obtained by modifying the already generated semantic representations (SQL program) through a few simple rules. For example, in Table~\ref{tab:edit_case}, after attaining the raw SQL query in turn one, the user is too ``lazy'' to declare the full intention and only convey the difference. This can be easily captured by a simple column insertion on the target SQL query. Similarly, in turn 3, the complete SQL can be obtained via a simple deletion of the \textsc{Where} clause based on turn 2, instead of generating the tedious long output. To achieve this, we thoroughly analyze the entire training set and summarize $14$ unit edit rules~(\ref{sec:definition}). Next, we propose an abstract syntax tree~(AST) comparison algorithm to automatically extract the chain of edition rules with the minimum length~(\ref{sec:extraction}). After that, we serialize and prepend those editions in the prompt before the output of each turn. Different serialization styles are analyzed~(\ref{sec:style}), including self-defined edit rules, python code and natural language description. And we find that the NL description performs the best on two benchmarks, SParc~\cite{yu-etal-2019-sparc} and CoSQL~\cite{yu-etal-2019-cosql}.

Our contributions can be summarized:


\begin{enumerate}
    \item We propose CoE-SQL to tackle the complex multi-turn text-to-SQL, which formalizes the SQL editions as a specific reasoning process. This method is more interpretable towards how LLM deals with context modeling to simulate human thinking.
    \item We provide the checklist of unit edit rules, and the corresponding tree comparison algorithm to automatically extract the edition chain by comparing two abstract syntax trees~(ASTs).
    \item We conduct comprehensive ablation study to analyze different CoE configurations and achieve state-of-the-art results with LLMs on the validation sets of two benchmarks SParC and CoSQL. It is also competitive to SOTA fine-tuned models.
\end{enumerate}

%% file: contents/related_work.tex
\section{Related Work}

\paragraph{Multi-turn text-to-SQL models} Before LLMs are applied in the multi-turn text-to-SQL task, researches mainly focus on building and fine-tuning specialized deep neural networks. ~\cite{zhang-etal-2019-editing} and ~\cite{wang2020tracking} use the previously generated SQL queries to improve the parsing accuracy. IGSQL~\cite{cai-wan-2020-igsql} utilizes the graph neural network to model database schema items in the conversational scenario. R$^2$SQL~\cite{hui2021dynamic} and HIE-SQL~\cite{zheng-etal-2022-hie} present a dynamic schema-linking graph which incorporates the current utterance, the previous preceding utterances, the database schema, and the last most recent SQL query. RASAT~\cite{qi-etal-2022-rasat} is a Transformer~\cite{vaswani2023attention} architecture augmented with relation-aware self-attention that could leverage a variety of relational structures while effectively inheriting the pre-trained parameters from the T5 model~\cite{raffel2023exploring}. RASAT employs the PICARD method~\cite{scholak-etal-2021-picard} which constrains the auto-regressive decoder by rejecting invalid tokens.

Despite the impressive results of specialized models, there are some unavoidable drawbacks. Creating and labeling a comprehensive text-to-SQL dataset requires a significant amount of resources and time. Additionally, training and refining the model is a laborious process that requires a lot of computing power.

\paragraph{In-context learning for text-to-SQL} Recent studies have explored the potential of LLMs for the text-to-SQL task, with \citet{rajkumar2022evaluating} using the zero-shot and few-shot learning setting to empirically evaluate the capabilities of LLMs such as GPT-3~\cite{brown2020language} and Codex~\cite{chen2021evaluating}. \citet{nan2023enhancing} focused on the strategy of exemplar selection, requiring an additional predictor to assess the difficulty of the SQL. DIN-SQL~\cite{pourreza2023dinsql} provides a more complex approach, decomposing the problem into several simpler sub-problems.

The above works merely employ LLMs on the single-turn text-to-SQL task. ACT-SQL~\cite{zhang2023actsql} generates the chain-of-thoughts automatically and extends its approach onto the multi-turn text-to-SQL task. ACT-SQL converts the multi-turn dataset into the single-turn one by rewriting and completing questions with context dependencies. However, ACT-SQL performs poorly under the multi-turn setup due to the error propagation occurring in the process of question rewriting. In contrast, our proposed CoE-SQL is an edit-based method which can directly utilize the context dependency instead of rewriting the question.

%% file: contents/methodology.tex
\section{Methodology}

In the few-shot in-context learning setting, the multi-turn text-to-SQL task can be formulated as
$$
R_n = \mathrm{LLM}(I, D, \mathcal{Q}_{\le n}, \mathcal{R}_{< n}, \mathcal{E}).
$$
$R_n$ represents the response to the current question created by LLMs. $I$ represents the instruction. $D$ represents the database schema. $\mathcal{Q} = [Q_1, Q_2, \cdots, Q_n]$ represents the entire context consisting of $n$ questions. $\mathcal{R} = [R_1, R_2, \cdots, R_{n - 1}]$ represents LLMs' responses to the previous questions. $\mathcal{E} = [E_1, E_2, \cdots, E_{|\mathcal{E}|}]$ is the list of $|\mathcal{E}|$ exemplars used in few-shot learning.


\subsection{Overview of CoE}

In the real-world scenario, users are more likely to start the conversation with a relatively simple question because they are unfamiliar with the detailed structure of the system. With the increasing the number of conversation turns, the user question and the corresponding SQL query will become more complex. It is more difficult and redundant for LLMs to generate a complex SQL query from scratch, since the entire thinking and logical reasoning process is generally intricate. On the contrary, generating the current SQL query by updating the previous one through a few editions is a better option.

In Section \ref{sec:definition}, we provide our definition of unit edit rules that can help edit the SQL query. In Section \ref{sec:extraction}, we explain how to extract the edition chain by comparing the two ASTs of two SQL queries. In Section \ref{sec:style}, we introduce the different styles of edition chains used in our work. And finally in Section \ref{sec:analysis}, we provide a simple method to help LLMs better analyze the edition process.


\subsection{Definition of Unit Edit Rules}
\label{sec:definition}

In order to edit a SQL query into another SQL query, we first define the set of unit edit rules. According to the different SQL components, we totally define 14 unit edit rules shown in Table \ref{tab:unit_edit_rule}. Taking the conversation instance in Table \ref{tab:edit_case} as an example, we can apply the unit edit rule EditSelectItem(-, market.District) to edit SQL 1 into SQL 2. We can also apply the EditWhereCondition(market.District = ``Alberta'', -) unit edit rule to edit SQL 2 into SQL 3.

\subsection{Extraction of Edition Chains}
\label{sec:extraction}

We use the few-shot learning method to activate LLMs' ability of utilizing our pre-defined unit edit rules. Therefore, we need to select exemplars from the training dataset and then extract the edition chains in each conversation. Since this work does not focus on selecting better exemplars, we use a simple exemplars selection strategy. We first randomly choose $k_d$ database schemas and then randomly choose $k_e$ dataset examples for each database schema. Thus, total $k_d \times k_e$ exemplars are put in the prompting text for the few-shot learning. In the following process, we need to extract the edition chains from these exemplars.

Assume that the dataset example consists of $n$ questions $[Q_1, Q_2, \cdots, Q_n]$ and $n$ corresponding SQL queries $[S_1, S_2, \cdots, S_n]$. Consider the $i$-th one as the current turn. Notably, the edition chain is determined by the difference between the current SQL query $S_i$ and the previous SQL query $S_j (j < i)$. A reasonable approach to extracting the edition chain is to compare the two ASTs.

\normalem
\begin{algorithm}
\caption{Extraction of Edition Chains}
\label{alg:extraction}
\SetKwInOut{Input}{Input}
\SetKwInOut{Output}{Output}
\Input{Previous tree node $t_\mathrm{old}$, current tree node $t_\mathrm{new}$}
\Output{Edition chain $C$}
$C \leftarrow \{\}$\;
\For{$f \ \mathrm{in \ get\_fields}(t_\mathrm{old})$}{
    $s_\mathrm{old} \leftarrow t_\mathrm{old}.\mathrm{get\_son}(f)$\;
    $s_\mathrm{new} \leftarrow t_\mathrm{new}.\mathrm{get\_son}(f)$\;
    \If{$s_\mathrm{old} \ \mathrm{is \ tree \ and} \ s_\mathrm{new} \ \mathrm{is \ tree}$}{
        $C_\mathrm{sub} \leftarrow \mathrm{ExtractCoE}(s_\mathrm{old}, s_\mathrm{new})$\;\tcp{recursively call the function}
        $C.\mathrm{update}(C_\mathrm{sub})$\;
    }
    \ElseIf{$s_\mathrm{old} \ \mathrm{is \ None \ and} \ s_\mathrm{new} \ \mathrm{is \ tree}$}{
        $e \leftarrow \mathrm{get\_add\_edition}(s_\mathrm{new}, f)$\;
        $C.\mathrm{add}(e)$\;
    }
    \ElseIf{$s_\mathrm{old} \ \mathrm{is \ tree \ and} \ s_\mathrm{new} \ \mathrm{is \ None}$}{
        $e \leftarrow \mathrm{get\_delete\_edition}(s_\mathrm{old}, f)$\;
        $C.\mathrm{add}(e)$\;
    }
    \ElseIf{$s_\mathrm{old} \neq s_\mathrm{new}$}{
        $e \leftarrow \mathrm{get\_change\_edition}(s_\mathrm{old}, s_\mathrm{new}, f)$\;
        $C.\mathrm{add}(e)$\;
    }
}
\Return{C}
\end{algorithm}
\ULforem

\begin{figure}[htbp]
    \centering
    \includegraphics[width=0.49\textwidth]{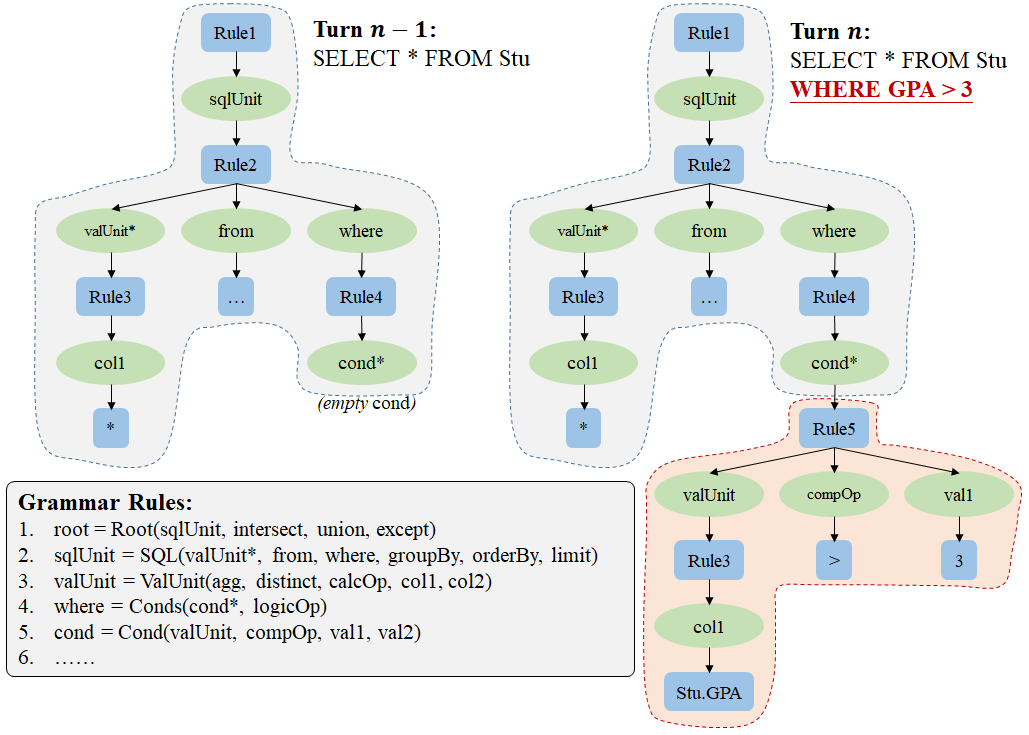}
    \caption{Comparison between two ASTs.}
    \label{fig:edit_ast}
\end{figure}

Figure \ref{fig:edit_ast} shows an example of a comparison between two ASTs. Notice that the FROM clause component is omitted in this figure. We compare each node pair in the two ASTs. Two nodes are considered equal iff they represent the same grammar rule and all of their child nodes are equal. The edition chain can be constructed according to the unequal part. By recognizing the grammar rules of the nodes in the unequal part, we can determine that the edition chain contains exactly one unit edit rule, i.e. EditWhereCondition(-, Stu.GPA > 3). The detailed procedure is outlined in Algorithm \ref{alg:extraction}. With the method of comparing two ASTs, we can construct the edition chains for every exemplar automatically. LLMs can learn the chain-of-editions method during the few-shot learning process.

In quite a few conversations, the current question $Q_i$ may not inherit from the exactly previous question $Q_{i - 1}$ but the more previous one $Q_j (j < i - 1)$. Also, the current question may start a new topic irrelevant to the original one. To tackle the above two problems, we enumerate all the previous $i - 1$ SQL queries and construct an edition chain $\mathrm{CoE}_{i, j}$ for each SQL pair $(S_i, S_j)$ where $j = 1, 2, \cdots, i - 1$. We eliminate those edition chains with lengths greater than $l_c$ (edition chains consisting of more than $l_c$ unit edit rules) where $l_c$ is a threshold. If the edition chain contains too many unit edit rules, we believe that the corresponding two questions are irrelevant. In that case, writing the current SQL query directly is more efficient than editing from the previous one. As for the left edition chains, we select the edition chain with the least number of tokens as the final edition chain.

\subsection{Style of Edition Chains}
\label{sec:style}

We attempt to use three different styles to put the edition chain in the prompting text. Table ~\ref{tab:coe_style} shows the detailed formats where the example chain-of-edition consists of two unit edit rules, namely EditSelectItem(*, COUNT(*)) and EditWhereCondition(-, Stu.GPA > 3).

\begin{table}[htbp]
\centering
\resizebox{0.49\textwidth}{!}{
\begin{tabular}{c|c}
\hline

\hline
\textbf{Style} & \textbf{Prompting Text} \\
\hline

\hline
Edit Rule & \makecell[l]{FROM clause: \\ - no change is needed \\ SELECT clause: \\ - EditSelectItem(*, COUNT(*)) \\ WHERE clause: \\ - EditWhereCondition(-, Stu.GPA > 3) \\ GROUP BY clause: \\ - no change is needed \\ ORDER BY clause: \\ - no change is needed \\ LIMIT clause: \\ - no change is needed \\ INTERSECT/UNION/EXCEPT: \\ - no change is needed} \\
\hline
Python Code & \makecell[l]{sql['select'].remove('*') \\ sql['select'].append('COUNT(*)') \\ sql['where']['conditions'].append('Stu.GPA > 3')} \\
\hline
Natural Language & \makecell[l]{FROM clause: \\ - no change is needed \\ SELECT clause: \\ - change * to COUNT(*) \\ WHERE clause: \\ - add WHERE condition Stu.GPA > 3 \\ GROUP BY clause: \\ - no change is needed \\ ORDER BY clause: \\ - no change is needed \\ LIMIT clause: \\ - no change is needed \\ INTERSECT/UNION/EXCEPT: \\ - no change is needed} \\
\hline

\hline
\end{tabular}
}
\caption{Three styles of serialization for CoE.}
\label{tab:coe_style}
\end{table}

With the edit rule style, we directly write our unit edit rules into the prompting text. With the Python code style, we regard the SQL query as a Python dict and convert unit edit rules into corresponding Python codes to update the Python dict. With the natural language style, unit edit rules are converted into plain texts which are closer to the corpus used in the LLMs' pretraining step.

\subsection{Analysis of Differences Between Pre- and Post-Questions}
\label{sec:analysis}

When preprocessing the exemplar, the edition chain can be extracted by comparing the different ASTs. However, when handling the test case, LLMs have to predict the edition chain by comparing the current question and the previous question. Thus we complete the question analysis for the exemplars. Given the current question and the previous question, we instruct LLMs to generate the difference between them. The analysis texts of each exemplar are also added into the prompting text, which can motivate LLMs to analyze the difference between pre- and post-questions. Notice that, this thought-before-edition trick is an adaptation of the thought-before-action idea in ReAct~\cite{react} framework.

%% file: contents/experiments.tex
\section{Experiments}

\subsection{Experiment Setup}

\paragraph{Models} We use the GPT-3.5-turbo-16k model to evaluate our proposed CoE-SQL. The CoE-SQL approach is based on self-defined edit operations, indicating that this method would not perform well if insufficient exemplars are provided. Only sufficient exemplars can cover most unit edit rules. Therefore, we expect LLMs to have a long context window. That's the reason we choose GPT-3.5-turbo-16k with a 16,385 tokens context window instead of GPT-3.5-turbo with a 4,096 tokens context window.

\paragraph{Hyperparameters} The temperature in the API of LLMs is set to 0, meaning that the greedy decoding strategy is being used. Text-to-SQL tasks require the model to produce SQL queries that adhere to strict grammar regulations. If the temperature is too high, the LLMs are likely to generate SQL queries that are invalid or not pertinent to the posed questions. Regarding the exemplars used for the few-shot learning, we set the number of database schemas $k_d$ to 4 and set the number of examples from each database schema $k_e$ to 4.

\paragraph{Datasets} We assess our proposed approach on SParC~\cite{yu-etal-2019-sparc} and CoSQL~\cite{yu-etal-2019-cosql}. SParC is composed of 4,298 coherent question sequences, including more than 12k individual questions and the related SQL queries. CoSQL has 10k+ annotated SQL queries. Each dialogue in CoSQL is designed to mimic a real-world situation, where a regular user is exploring the database and an expert is retrieving answers with SQL. They also provide an evaluation script that divides SQL queries into four difficulty levels (easy, medium, hard, and extra).

\paragraph{Evaluation metrics} We assess the performance of our approach using three commonly used evaluation metrics for the text-to-SQL task: exact match accuracy (EM), execution accuracy (EX), and test-suite accuracy (TS). EM requires that each component of the predicted SQL is the same as the corresponding component of the gold SQL, disregarding the values in the query. EX evaluates the correctness of the execution result of the predicted SQL, which is usually more precise than EM. TS also evaluates the execution result, but requires the result to be correct across multiple database instances per database schema\footnote{\url{https://github.com/taoyds/test-suite-sql-eval}}.

Since we are evaluating LLMs' performances in the multi-turn text-to-SQL task, question match accuracy (QM) and interaction match accuracy (IM) need to be considered respectively. QM is 1 if the predicted SQL query for the single question is correct, and IM is 1 if all the predicted SQL queries in the context are correct.

\subsection{Main Results}

\begin{table*}[htbp]
\centering
\begin{tabular}{c|ccc|ccc}
\hline

\hline
\multirow{2}*{\textbf{Fine-tuned Model}} & \multicolumn{3}{c|}{\textbf{QM}} & \multicolumn{3}{c}{\textbf{IM}} \\
\cline{2-7}
~ & \textbf{EM $\uparrow$} & \textbf{EX $\uparrow$} & \textbf{TS $\uparrow$} & \textbf{EM $\uparrow$} & \textbf{EX $\uparrow$} & \textbf{TS $\uparrow$} \\
\hline

\hline
GAZP+BERT~\cite{zhong-etal-2020-grounded} & 48.9 & 47.8 & - & - & - & - \\
HIE-SQL+GraPPa~\cite{zheng-etal-2022-hie} & 64.7 & - & - & 45.0 & - & - \\
RASAT+PICARD~\cite{qi-etal-2022-rasat} & \textbf{67.7} & \textbf{73.3} & - & \textbf{49.1} & \textbf{54.0} & - \\
\hline

\hline
\textbf{In-Context Learning Approach} & \multicolumn{6}{c}{~} \\
\hline

\hline
ACT-SQL~\cite{zhang2023actsql} & 51.0 & 63.8 & 56.9 & 24.4 & 38.9 & 29.6 \\
Baseline (Ours) & 50.0 & 67.0 & 59.5 & 30.8 & 46.7 & 37.9 \\
\textbf{CoE-SQL (Ours)} & \textbf{56.0} & \textbf{70.3} & \textbf{63.3} & \textbf{36.5} & \textbf{50.5} & \textbf{41.9} \\
\hline

\hline
\end{tabular}
\caption{Performances of CoE-SQL and other previous works on SParC dev set.}
\label{tab:sparc_dev}
\end{table*}

\begin{table*}[htbp]
\centering
\begin{tabular}{c|ccc|ccc}
\hline

\hline
\multirow{2}*{\textbf{Fine-tuned Model}} & \multicolumn{3}{c|}{\textbf{QM}} & \multicolumn{3}{c}{\textbf{IM}} \\
\cline{2-7}
~ & \textbf{EM $\uparrow$} & \textbf{EX $\uparrow$} & \textbf{TS $\uparrow$} & \textbf{EM $\uparrow$} & \textbf{EX $\uparrow$} & \textbf{TS $\uparrow$} \\
\hline

\hline
GAZP+BERT~\cite{zhong-etal-2020-grounded} & 42.0 & 38.8 & - & - & - & - \\
HIE-SQL+GraPPa~\cite{zheng-etal-2022-hie} & 56.4 & - & - & \textbf{28.7} & - & - \\
RASAT+PICARD~\cite{qi-etal-2022-rasat} & \textbf{58.8} & \textbf{67.0} & - & 27.0 & \textbf{39.6} & - \\
\hline

\hline
\textbf{In-Context Learning Approach} & \multicolumn{6}{c}{~} \\
\hline

\hline
ACT-SQL~\cite{zhang2023actsql} & 46.0 & 63.7 & 55.2 & 13.3 & 30.7 & 21.5 \\
Baseline (Ours) & 47.8 & 69.4 & 58.5 & 20.1 & 38.9 & 27.6 \\
\textbf{CoE-SQL (Ours)} & \textbf{52.4} & \textbf{69.6} & \textbf{60.6} & \textbf{23.9} & \textbf{39.6} & \textbf{30.4} \\
\hline

\hline
\end{tabular}
\caption{Performances of CoE-SQL and other previous works on CoSQL dev set.}
\label{tab:cosql_dev}
\end{table*}

In our main experiments, we choose the natural language style for the edition chain. We set the maximum length of the edition chain $l_c$ to 4 when testing on SParC and 3 when testing on CoSQL. Table \ref{tab:sparc_dev} and Table \ref{tab:cosql_dev} show the performance of our proposed CoE-SQL and other previous works on the dev sets of SParC and CoSQL respectively.

Notably, when comparing in-context learning approaches with fine-tuned models, the EM evaluation metric is not that worthy to be paid attention to. Fine-tuned models can learn the dataset feature from the training set. These models are more likely to generate the SQL query with the same structure as the gold SQL query and thus can achieve higher EM scores. On the contrary, LLMs tend to write the SQL query based on their original knowledge learning in the pretraining phase. Only a few exemplars from the training dataset cannot provide sufficient information about the dataset feature. Therefore, LLMs are more likely to generate the SQL query with the accurate semantic and logic and the correct execution result. In general, we would like to mainly focus on the EX and TS evaluation metrics in the following discussion. Most fine-tuned models only provide their EM scores. We compare our method with the GAZP and the RASAT methods because these two models provide their EX scores, where RASAT is the SOTA one.

Compared with fine-tuned models, our proposed CoE-SQL approach achieves a 70.3\% EX(QM) score and a 50.5\% EX(IM) score on SParC dev set, which has surpassed the GAZP + BERT model~\cite{zhong-etal-2020-grounded} a lot and has been comparable to the RASAT + PICARD model~\cite{qi-etal-2022-rasat}. CoE-SQL even achieves the highest EX(QM) score on CoSQL dev set. The experiment result proves that LLMs have possessed the strong ability for handling the complex multi-turn text-to-SQL task. Using the GPT-3.5-turbo-16k LLM, the CoE-SQL approach can perform almost as well as the previous best fine-tuned model (with EX score). We believe that our CoE-SQL can achieve a better performance if larger LLMs (e.g. GPT-4) are applied.

Furthermore, the CoE-SQL approach achieves the highest EM, EX, and TS scores among the existing in-context learning methods. The ACT-SQL method converts the multi-turn dataset into the single-turn one by rewriting and completing the questions with context dependencies. Comparing our simple baseline method and the ACT-SQL method, we can conclude that paraphrasing the multi-turn dataset with LLMs is not a good choice. It performs even worse than the baseline method. Based on the edit operations, CoE-SQL performs much better than the baseline method which merely takes the original database schema and questions as the LLMs' input. This indicates that editing SQL queries is an effective method in the multi-turn text-to-SQL task.

\begin{table*}[htbp]
\centering
\begin{tabular}{c|cc|cc}
\hline

\hline
\multirow{2}*{\textbf{Fine-tuned Model}} & \multicolumn{2}{c|}{\textbf{SParC}} & \multicolumn{2}{c}{\textbf{CoSQL}} \\
\cline{2-5}
~ & \textbf{EX (QM) $\uparrow$} & \textbf{EX (IM) $\uparrow$} & \textbf{EX (QM) $\uparrow$} & \textbf{EX (IM) $\uparrow$} \\
\hline

\hline
GAZP+BERT~\cite{zhong-etal-2020-grounded} & 44.6 & 19.7 & 35.9 & 8.4 \\
RASAT+PICARD~\cite{qi-etal-2022-rasat} & 74.0 & \textbf{52.6} & 66.3 & 37.4 \\
\hline

\hline
\textbf{In-Context Learning Approach} & \multicolumn{4}{c}{~} \\
\hline

\hline
\textbf{CoE-SQL (Ours)} & \textbf{74.1} & 51.9 & \textbf{71.1} & \textbf{42.9} \\
\hline

\hline
\end{tabular}
\caption{Performances of CoE-SQL and other previous works on SParC and CoSQL test sets.}
\label{tab:sparc_cosql_test}
\end{table*}

We also complete the experiments on SParC and CoSQL test sets. The results of the experiment are shown in Table \ref{tab:sparc_cosql_test}. Under the EX metric, our CoE-SQL approach keeps the good performance and even surpasses the SOTA fine-tuned model on the test sets. This also reveals that in-context learning approaches would not suffer the performance gap between the dev set and the test set.

\paragraph{Case study} Table \ref{tab:case_study} shows an instance from the SParC dev set. The condition ``employees under age 30'' is mentioned in the first question. The baseline method successfully keeps this condition in the second query but forgets it in the final query. However, due to the chain-of-edition mechanism, the CoE-SQL method retains this condition from the beginning to the end. More detailed case studies can be found in Appendix \ref{app:b}.

\begin{table}[htbp]
\centering
\resizebox{0.49\textwidth}{!}{
\begin{tabular}{p{\columnwidth}}
\hline

\hline
\textbf{Turn 1} \\
\textbf{Question:} {\it Find all employees who are under age 30.} \\
\textbf{Baseline:} SELECT * FROM employee WHERE age < 30 \textcolor[RGB]{48,128,20}{\large\checkmark} \\
\textbf{CoE-SQL:} SELECT * FROM employee WHERE age < 30 \textcolor[RGB]{48,128,20}{\large\checkmark} \\
\hdashline
\textbf{Turn 2} \\
\textbf{Question:} {\it Which cities did they come from?} \\
\textbf{Baseline:} SELECT DISTINCT city FROM employee WHERE age < 30 \textcolor[RGB]{48,128,20}{\large\checkmark} \\
\textbf{CoE-SQL:} SELECT City FROM employee WHERE age < 30 \textcolor[RGB]{48,128,20}{\large\checkmark} \\
\hdashline
\textbf{Turn 3} \\
\textbf{Question:} {\it Show the cities from which more than one employee originated.} \\
\textbf{Baseline:} SELECT city FROM employee GROUP BY city HAVING count(*) > 1 \textcolor[RGB]{176,23,31}{\scriptsize\XSolidBold} \\
\textbf{CoE-SQL:} SELECT City FROM employee \textcolor[RGB]{0,112,192}{WHERE age < 30} GROUP BY City HAVING COUNT(*) > 1 \textcolor[RGB]{48,128,20}{\large\checkmark} \\
\hline

\hline
\end{tabular}
}
\caption{Case study from the SParC dev set.}
\label{tab:case_study}
\end{table}

\subsection{Ablation Study}

In this section, we show the experiment results of the ablation study to find out the best setting of CoE-SQL. We only report the TS score since it is most suitable for the in-context learning method.

\paragraph{Simple CoT baseline} Besides the most trivial baseline method, we attempt an advanced approach based on the simple CoT method. We manually label the few-shot exemplars using the question explanation mentioned in~\cite{chen2023teaching}. Table \ref{tab:as_cot} shows the result, which proves that the simple CoT method can be beneficial for LLMs compared with the most trivial baseline. It is also significant that our CoE-based approach is more suitable for the multi-turn text-to-SQL task.

\begin{table}[htbp]
\centering
\begin{tabular}{c|cc}
\hline

\hline
\textbf{Approach} & \textbf{TS(QM) $\uparrow$} & \textbf{TS(IM) $\uparrow$} \\
\hline

\hline
Baseline & 59.5 & 37.9 \\
Simple CoT & 61.8 & 39.8 \\
CoE-SQL & \textbf{63.3} & \textbf{41.9} \\
\hline

\hline
\end{tabular}
\caption{Performances of CoE-SQL and baseline methods on SParC dev set.}
\label{tab:as_cot}
\end{table}

\paragraph{Style of edition chains} Table \ref{tab:as_coe_style} shows the performance of CoE-SQL on SParC dev set influenced by three styles of prompting text for chain-of-editions mentioned in Section \ref{sec:style}. The experiment result proves that the natural language style is the most suitable one. This is because LLMs like GPT models are mostly trained with natural language corpuses. The chain-of-editions style with the edit rule performs relatively poor, since our unit edit rules are self-designed and very unlikely to appear in the pretraining corpus. Although LLMs must have seen many Python codes during pretraining, the Python-code style still receives bad scores. We believe that this is because the Python codes generated by LLMs are used to update the Python dict that represents the SQL query. The structure of this Python dict is complex and unfamiliar for LLMs, though we have provided the structure in the instruction and exemplars.

\begin{table}[htbp]
\centering
\resizebox{0.49\textwidth}{!}{
\begin{tabular}{c|cc}
\hline

\hline
\textbf{Style} & \textbf{TS(QM) $\uparrow$} & \textbf{TS(IM) $\uparrow$} \\
\hline

\hline
Edit Rule & 61.2 & 40.5 \\
Python Code & 58.6 & 37.9 \\
Natural Language & \textbf{63.3} & \textbf{41.9} \\
\hline

\hline
\end{tabular}
}
\caption{CoE-SQL performance on SParC dev set influenced by three styles of prompting text for chain-of-editions.}
\label{tab:as_coe_style}
\end{table}

When using the Python-code style, the prompting text mainly consists of three parts, i.e. the Python code that represents the edit rule, the Python dict that represents the SQL components, and the current SQL query. Thus we complete more ablation studies about these LLMs-generated parts. Table \ref{tab:as_python} shows the result. First, we change the order of the Python dict and the SQL query in the exemplar. Second, we complete the post-processing according to different parts which are marked with ``*'' symbols in the table. When post-processing with the code, we run the LLM-generated code to update the dict and get the SQL by parsing the updated dict. When post-processing with the dict, we directly get the SQL by parsing the dict generated by LLMs. The experiment shows that appending the dict after the SQL is generally a better choice. The Python codes generated by LLMs are not that reliable. It is better to directly use the Python dict or the SQL itself.

\begin{table}[htbp]
\centering
\begin{tabular}{c|cc}
\hline

\hline
\textbf{Python-Code Style} & \textbf{TS(QM) $\uparrow$} & \textbf{TS(IM) $\uparrow$} \\
\hline

\hline
code*+dict+SQL & 51.8 & 28.7 \\
code+dict*+SQL & 57.2 & 34.8 \\
code+dict+SQL* & 58.3 & 36.3 \\
code*+SQL+dict & 52.5 & 29.1 \\
code+SQL+dict* & \textbf{58.6} & \textbf{37.9} \\
code+SQL*+dict & 56.9 & 36.0 \\
\hline

\hline
\end{tabular}
\caption{CoE-SQL performance on SParC dev set influenced by different Python-code styles of prompting text for chain-of-editions. Different python-code styles indicate that the three components code, dict, and SQL are put in the prompting text in different orders. The ``*'' mark means that we use this specific component to complete the post-processing procedure.}
\label{tab:as_python}
\end{table}

When using the natural language style, we add ``no change is needed'' in the prompt if the clause is not edited as shown in Table \ref{tab:coe_style}. The experiment result in Table \ref{tab:as_no_change} proves that this prompting text is necessary. Through adding this special sentence, the CoE prompting text in the exemplars can be regular and normalized, since all the components and clauses can be mentioned in the context. LLMs are generally better at receiving and handling regular and normalized contexts.

\begin{table}[htbp]
\centering
\resizebox{0.49\textwidth}{!}{
\begin{tabular}{c|cc}
\hline

\hline
\textbf{Natural Language Style} & \textbf{TS(QM) $\uparrow$} & \textbf{TS(IM) $\uparrow$} \\
\hline

\hline
Natural Language & \textbf{63.3} & \textbf{41.9} \\
w/o ``no change is needed'' & 62.8 & 41.0 \\
\hline

\hline
\end{tabular}
}
\caption{CoE-SQL performance on SParC dev set influenced by the ``no change is needed'' prompting text.}
\label{tab:as_no_change}
\end{table}

\paragraph{Analysis of differences between pre- and post-Questions} Table \ref{tab:as_analysis} proves that the question analysis mentioned in Section \ref{sec:analysis} is effective in our approach. The analysis of differences between the current question and the previous one is beneficial for LLMs to think more about the possible edit rules. Without the analysis, the TS(QM) score and the TS(IM) score both drop about 2\%.

\begin{table}[htbp]
\centering
\begin{tabular}{c|cc}
\hline

\hline
\textbf{Method} & \textbf{TS(QM) $\uparrow$} & \textbf{TS(IM) $\uparrow$} \\
\hline

\hline
CoE-SQL & \textbf{63.3} & \textbf{41.9} \\
w/o analysis & 61.3 & 39.8 \\
\hline

\hline
\end{tabular}
\caption{CoE-SQL performance on SParC dev set influenced by the question analysis.}
\label{tab:as_analysis}
\end{table}

\begin{figure}[htbp]
    \centering
    \includegraphics[width=0.49\textwidth]{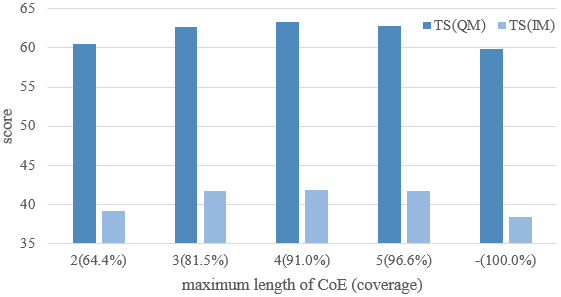}
    \caption{CoE-SQL performance on SParC dev set influenced by the maximum length of the edition chain.}
    \label{fig:as_max_len}
\end{figure}

\paragraph{Coverage of edition chain} As mentioned in Section \ref{sec:extraction}, if an edition chain is too long, LLMs would generate the SQL query directly instead of using the edit-based method. We can control the coverage of the edition chain on the training dataset by changing the maximum length of the edition chain $l_c$. Figure \ref{fig:as_max_len} shows the performances on SParC dev set influenced by $l_c$. According to the experiment result, we set $l_c$ to 4 in our main experiment. If $l_c$ is too small, LLMs would be more likely to directly generate the SQL query without using the edit-based method. If $l_c$ is too large, LLMs would always edit the SQL query although the CoE-SQL approach may not be suitable for the current testing case.





We also try another way to control the chain-of-edition coverage in the training dataset. We remove the relatively complex unit edit rules and only retain the commonly used ones. Therefore, LLMs would not be forced to edit the SQL query if the SQL query contains complex clauses. Table \ref{tab:as_unit_edit_rule} shows the experiment results with the different definitions of unit edit rules. In the first row, we only keep the unit edit rules involving the SELECT clause and the WHERE clause. In the second row, we add the unit edit rules that are relative to the FROM clause. In the third row, we do not set any limitations. The experiment result indicates that this method is not that effective. With the change of the unit edit rules, the LLMs' performance does not change a lot. This explains why we use the maximum length of the edition chain to limit the edit-based approach.

\begin{table}[htbp]
\centering
\resizebox{0.49\textwidth}{!}{
\begin{tabular}{c|cc}
\hline

\hline
\textbf{Unit Edit Rules (Coverage)} & \textbf{TS(QM) $\uparrow$} & \textbf{TS(IM) $\uparrow$} \\
\hline

\hline
S + W (48.1\%) & 60.0 & 38.9 \\
F + S + W (65.9\%) & 60.3 & 38.2 \\
- (100.0\%) & 59.9 & 38.4 \\
\hline

\hline
\end{tabular}
}
\caption{CoE-SQL performance on SParC dev set influenced by the definition of unit edit rules. F represents the unit edit rules about the FROM clause. S represents the unit edit rules about the SELECT clause. W represents the unit edit rules about the WHERE clause.}
\label{tab:as_unit_edit_rule}
\end{table}

\subsection{Discussion}

We discuss the effectiveness of our CoE-SQL method under the SParC dev set. There are totally 422 interactions and 1,203 SQL queries in the SParC dev set. Among all 1,203 SQL results written by LLMs, 484 SQL queries are generated directly without using edit rules. Moreover, the first SQL in each interaction is always generated directly. Ignoring the first SQL in each interaction, the probability of generating SQL directly is merely $\frac{484 - 422}{1203 - 422} \times 100\% = 7.9\%$. This indicates that LLMs effectively utilize our CoE-based approach to handle most complex testing cases. The CoE-based approach takes a significant role in the prediction process.

%% file: contents/conclusion.tex
\section{Conclusion}

We propose our CoE-SQL approach for the multi-turn text-to-SQL task based on editing the previous SQL query to the current SQL query. We explore the definition and the style of our unit edit rules. We also provide the method to extract the edition chain by comparing two ASTs of two different SQL queries. Furthermore, our proposed CoE framework follows the human thinking process. The experiment results demonstrate that our approach achieves the best performances on the SParC and CoSQL dev set among existing in-context learning approaches and is also comparable to the SOTA fine-tuned model. We complete some ablation studies and prove the effectiveness of various components in CoE-SQL.

%% file: contents/limitations.tex
\section*{Limitations}

There are some limitations in our work. First, we mainly concentrate on investigating the effectiveness of the edition chain. We do not explore anything about the exemplar selection strategy which can influence LLMs' performances a lot. Second, we cannot ensure that we have thoroughly optimized the CoE-SQL approach. There still may exist some methods for the optimization. Third, our approach does not surpass the previous SOTA fine-tuned model on the SParC and CoSQL dev sets under some evaluation metrics. These are all difficult tasks that need to be addressed in the future work.

%% file: contents/acknowledgements.tex
\section*{Acknowledgements}

This work is funded by the China NSFC Projects (62106142, U23B2057 and 62120106006) and Shanghai Municipal Science and Technology Major Project (2021SHZDZX0102).

%% file: contents/appendix_a.tex
\section{Unit Edit Rules}

\begin{table*}[htbp]
\centering
\resizebox{0.99\textwidth}{!}{
\begin{tabular}{c|c}
\hline

\hline
\textbf{Unit Edit Rule} & \textbf{Explanation} \\
\hline

\hline
EditSelectItem(oldItem, newItem) & \makecell[c]{Replace oldItem with newItem in the SELECT clause. \\ Add newItem into the SELECT clause if oldItem is ``-''. \\ Delete oldItem from the SELECT clause if newItem is ``-''.} \\
\hline
EditFromTable(oldTable, newTable) & \makecell[c]{Replace oldTable with newTable in the FROM clause. \\ Add newTable into the FROM clause if oldTable is ``-''. \\ Delete oldTable from the FROM clause if newTable is ``-''.} \\
\hline
EditNestedFromClause(SQL) & \makecell[c]{Edit the nested FROM clause with SQL. \\ Delete the nested FROM clause if SQL is ``-''.} \\
\hline
EditJoinCondition(oldCondition, newCondition) & \makecell[c]{Replace oldCondition with newCondition in the ON clause. \\ Add newCondition into the ON clause if oldCondition is ``-''. \\ Delete oldCondition from the ON clause if newCondition is ``-''.} \\
\hline
EditJoinLogicalOperator(and/or) & Edit the logical operator in the ON clause. \\
\hline
EditWhereCondition(oldCondition, newCondition) & \makecell[c]{Replace oldCondition with newCondition in the WHERE clause. \\ Add newCondition into the WHERE clause if oldCondition is ``-''. \\ Delete oldCondition from the WHERE clause if newCondition is ``-''.} \\
\hline
EditWhereLogicalOperator(and/or) & Edit the logical operator in the WHERE clause. \\
\hline
EditGroupByColumn(oldColumn, newColumn) & \makecell[c]{Replace oldColumn with newColumn in the GROUP BY clause. \\ Add newColumn into the GROUP BY clause if oldColumn is ``-''. \\ Delete oldColumn from the GROUP BY clause if newColumn is ``-''.} \\
\hline
EditHavingCondition(oldCondition, newCondition) & \makecell[c]{Replace oldCondition with newCondition in the HAVING clause. \\ Add newCondition into the HAVING clause if oldCondition is ``-''. \\ Delete oldCondition from the HAVING clause if newCondition is ``-''.} \\
\hline
EditHavingLogicalOperator(and/or) & Edit the logical operator in the HAVING clause. \\
\hline
EditOrderByItem(oldItem, newItem) & \makecell[c]{Replace oldItem with newItem in the ORDER BY clause. \\ Add newItem into the ORDER BY clause if oldItem is ``-''. \\ Delete oldItem from the ORDER BY clause if newItem is ``-''.} \\
\hline
EditOrder(asc/desc) & Edit the order in the ORDER BY clause. \\
\hline
EditLimit(oldLimit, newLimit) & \makecell[c]{Replace oldLimit with newLimit in the LIMIT clause. \\ Add newLimit into the LIMIT clause if oldLimit is ``-''. \\ Delete oldLimit from the LIMIT clause if newLimit is ``-''.} \\
\hline
EditIUE(intersect/union/except, left/right, SQL) & \makecell[c]{Append SQL to the left/right side of the previous SQL with intersect/union/except keyword. \\ Delete the left/right side of the previous SQL with intersect/union/except keyword if SQL is ``-''.} \\
\hline

\hline
\end{tabular}
}
\caption{All 14 defined unit edit rules.}
\label{tab:unit_edit_rule}
\end{table*}

%% file: contents/appendix_b.tex
\section{Detailed Experiment Results}
\label{app:b}

Table \ref{tab:detailed_sparc} and Table \ref{tab:detailed_cosql} show the detailed performances of ACT-SQL~\cite{zhang2023actsql}, our baseline, and our CoE-SQL on the SParC and CoSQL dev sets respectively. All the results in these two tables are TS scores. The results prove that the CoE-SQL approach performs better than the previous methods with different difficulty levels and numbers of conversational turns.

\begin{table*}[htbp]
\centering
\resizebox{0.99\textwidth}{!}{
\begin{tabular}{c|cc|cccc|ccccc}
\hline

\hline
\textbf{Approach} & \textbf{TS(QM)} & \textbf{TS(IM)} & \textbf{Easy} & \textbf{Medium} & \textbf{Hard} & \textbf{Extra} & \textbf{Turn 1} & \textbf{Turn 2} & \textbf{Turn 3} & \textbf{Turn 4} & \textbf{Turn > 4} \\
\hline

\hline
ACT-SQL~\cite{zhang2023actsql} & 56.9 & 29.6 & 77.2 & 51.9 & 40.0 & 17.9 & 72.5 & 55.7 & 41.5 & 34.1 & 100.0 \\
Baseline (Ours) & 59.5 & 37.9 & 77.6 & 58.5 & 42.8 & 15.7 & 69.2 & 59.5 & 49.6 & 43.2 & 100.0 \\
\textbf{CoE-SQL (Ours)} & 63.3 & 41.9 & 81.8 & 59.9 & 46.2 & 26.9 & 76.1 & 60.0 & 53.3 & 48.9 & 100.0 \\
\hline

\hline
\end{tabular}
}
\caption{Detailed performances on the SParC dev set.}
\label{tab:detailed_sparc}
\end{table*}

\begin{table*}[htbp]
\centering
\resizebox{0.99\textwidth}{!}{
\begin{tabular}{c|cc|cccc|ccccc}
\hline

\hline
\textbf{Approach} & \textbf{TS(QM)} & \textbf{TS(IM)} & \textbf{Easy} & \textbf{Medium} & \textbf{Hard} & \textbf{Extra} & \textbf{Turn 1} & \textbf{Turn 2} & \textbf{Turn 3} & \textbf{Turn 4} & \textbf{Turn > 4} \\
\hline

\hline
ACT-SQL~\cite{zhang2023actsql} & 55.2 & 21.5 & 76.5 & 46.6 & 42.3 & 17.8 & 65.2 & 54.7 & 50.8 & 48.2 & 42.3 \\
Baseline (Ours) & 58.5 & 27.6 & 79.9 & 48.7 & 46.6 & 22.4 & 66.6 & 58.2 & 55.3 & 56.1 & 40.8 \\
\textbf{CoE-SQL (Ours)} & 60.6 & 30.4 & 79.1 & 54.7 & 50.3 & 21.5 & 67.2 & 61.8 & 57.0 & 56.1 & 47.9 \\
\hline

\hline
\end{tabular}
}
\caption{Detailed performances on the CoSQL dev set.}
\label{tab:detailed_cosql}
\end{table*}

Table \ref{tab:detailed_case_studies} shows two testing cases from the SParC dev set. The complete LLMs' responses are shown in the table. In the first case, CoE-SQL successfully retains the WHERE condition while the trivial baseline method forgets this key information during the conversation. The second case demonstrates that the CoE-SQL approach can also assist LLMs to handle complex logical reasoning processes such as the JOIN operation for multi-tables and the GROUP BY operation in this case.

\begin{longtable}{p{\columnwidth}}
\hline

\hline
\textbf{Turn 1} \\
\textbf{Question:} {\it Find all employees who are under age 30.} \\
\textbf{Baseline:} SELECT * FROM employee WHERE age < 30 \textcolor[RGB]{48,128,20}{\large\checkmark} \\
\textbf{CoE-SQL:} \\
Let's think step by step. \\
SQL 17-1 can be written directly instead of being edited from previous SQL. \\
So SQL 17-1 is: \\
SELECT * FROM employee WHERE age < 30 \textcolor[RGB]{48,128,20}{\large\checkmark} \\
\hdashline
\textbf{Turn 2} \\
\textbf{Question:} {\it Which cities did they come from?} \\
\textbf{Baseline:} SELECT DISTINCT city FROM employee WHERE age < 30 \textcolor[RGB]{48,128,20}{\large\checkmark} \\
\textbf{CoE-SQL:} \\
Let's think step by step. \\
SQL 17-2 can be edited from SQL 17-1. \\
The previous question asked for all employees who are under the age of 30, while the current question asks for the cities they come from. \\
Therefore, following edit operations are used: \\
FROM clause: \\
- no change is needed \\
SELECT clause: \\
- change * to employee.City \\
WHERE clause: \\
- no change is needed \\
GROUP BY clause: \\
- no change is needed \\
ORDER BY clause: \\
- no change is needed \\
LIMIT clause: \\
- no change is needed \\
INTERSECT/UNION/EXCEPT: \\
- no change is needed \\
So SQL 17-2 is: \\
SELECT City FROM employee WHERE age < 30 \textcolor[RGB]{48,128,20}{\large\checkmark} \\
\hdashline
\textbf{Turn 3} \\
\textbf{Question:} {\it Show the cities from which more than one employee originated.} \\
\textbf{Baseline:} SELECT city FROM employee GROUP BY city HAVING count(*) > 1 \textcolor[RGB]{176,23,31}{\scriptsize\XSolidBold} \\
\textbf{CoE-SQL:} \\
Let's think step by step. \\
SQL 17-3 can be edited from SQL 17-2. \\
The previous question asked for the cities from which the employees originated, while the current question asks for the cities from which more than one employee originated. \\
Therefore, following edit operations are used: \\
FROM clause: \\
- no change is needed \\
SELECT clause: \\
- no change is needed \\
WHERE clause: \\
- no change is needed \\
GROUP BY clause: \\
- add HAVING condition COUNT(*) > 1 \\
ORDER BY clause: \\
- no change is needed \\
LIMIT clause: \\
- no change is needed \\
INTERSECT/UNION/EXCEPT: \\
- no change is needed \\
So SQL 17-3 is: \\
SELECT City FROM employee \textcolor[RGB]{0,112,192}{WHERE age < 30} GROUP BY City HAVING COUNT(*) > 1 \textcolor[RGB]{48,128,20}{\large\checkmark} \\
\hline
\textbf{Turn 1} \\
\textbf{Question:} {\it What are all the airlines?} \\
\textbf{Baseline:} SELECT Airline FROM airlines \textcolor[RGB]{48,128,20}{\large\checkmark} \\
\textbf{CoE-SQL:} \\
Let's think step by step. \\
SQL 17-1 can be written directly instead of being edited from previous SQL. \\
So SQL 17-1 is: \\
SELECT Airline FROM airlines \textcolor[RGB]{48,128,20}{\large\checkmark} \\
\hdashline
\textbf{Turn 2} \\
\textbf{Question:} {\it Which ones have more than 10 flights?} \\
\textbf{Baseline:} SELECT Airline FROM flights GROUP BY Airline HAVING count(*) > 10 \textcolor[RGB]{176,23,31}{\scriptsize\XSolidBold} \\
\textbf{CoE-SQL}: \\
Let's think step by step. \\
SQL 17-2 can be edited from SQL 17-1. \\
The previous question asked for all the airlines, while the current question asks for the airlines that have more than 10 flights. \\
Therefore, following edit operations are used: \\
FROM clause: \\
- add table flights \\
- add JOIN condition airlines.uid = flights.Airline \\
SELECT clause: \\
- no change is needed \\
WHERE clause: \\
- add WHERE condition COUNT(*) > 10 \\
GROUP BY clause: \\
- add column airlines.Airline \\
ORDER BY clause: \\
- no change is needed \\
LIMIT clause: \\
- no change is needed \\
INTERSECT/UNION/EXCEPT: \\
- no change is needed \\
So SQL 17-2 is: \\
SELECT airlines.Airline FROM \textcolor[RGB]{0,112,192}{airlines JOIN flights ON airlines.uid = flights.Airline} GROUP BY airlines.Airline HAVING COUNT(*) > 10 \textcolor[RGB]{48,128,20}{\large\checkmark} \\
\hline

\hline
\caption{Detailed case studies from the SParC dev set.}
\label{tab:detailed_case_studies}
\end{longtable}

%% file: contents/appendix_c.tex
\section{Experiment Results with Open-Source LLM}

\begin{table*}[htbp]
\centering
\begin{tabular}{c|ccc|ccc}
\hline

\hline
\multirow{2}*{\textbf{Approach}} & \multicolumn{3}{c|}{\textbf{QM}} & \multicolumn{3}{c}{\textbf{IM}} \\
\cline{2-7}
~ & \textbf{EM $\uparrow$} & \textbf{EX $\uparrow$} & \textbf{TS $\uparrow$} & \textbf{EM $\uparrow$} & \textbf{EX $\uparrow$} & \textbf{TS $\uparrow$} \\
\hline

\hline
CoE-SQL + lmsys/vicuna-7b-v1.5-16k & 23.7 & 36.0 & 29.9 & 11.4 & 19.4 & 14.5 \\
CoE-SQL + lmsys/vicuna-13b-v1.5-16k & 21.1 & 35.5 & 30.4 & 6.4 & 14.7 & 11.1 \\
\hline

\hline
\end{tabular}
\caption{Performances of CoE-SQL with open-source LLM on SParC dev set.}
\label{tab:open_source}
\end{table*}

We attempt to complete the experiment using the open-source LLM on the SParC dev set. Table \ref{tab:open_source} shows the result. We use the Vicuna LLM in our experiment. We find that open-source LLM's few-shot learning abilities are far inferior than GPT-series LLMs.

%% file: contents/appendix_d.tex
\section{Prompts}

In this section, we show our prompt engineering with an example from the SParC train set.

\subsection{Baseline}

role: system \\
content: \\
Given the database schema, you need to translate the question into the SQL query. \\
\\
role: user \\
content: \\
Database schema: \\
create table party ( \\
\hspace*{2em} Party\_ID number, \\
\hspace*{2em} Party\_Theme text, \\
\hspace*{2em} Location text, \\
\hspace*{2em} First\_year text, \\
\hspace*{2em} Last\_year text, \\
\hspace*{2em} Number\_of\_hosts number, \\
\hspace*{2em} primary key (Party\_ID) \\
) \\
/* \\
3 example rows from table party: \\
Party\_ID \hspace*{2em} Party\_Theme \hspace*{2em} Location \hspace*{2em} First\_year \hspace*{2em} Last\_year \hspace*{2em} Number\_of\_hosts \\
1 \hspace*{2em} Spring \hspace*{2em} Hemkade 48 Zaandam \hspace*{2em} 2000 \hspace*{2em} 2001 \hspace*{2em} 5 \\
2 \hspace*{2em} Club Q-BASE \hspace*{2em} Hemkade 48 Zaandam \hspace*{2em} 2002 \hspace*{2em} 2002 \hspace*{2em} 23 \\
3 \hspace*{2em} Hardcore Resurrection \hspace*{2em} Various locations \hspace*{2em} 2000 \hspace*{2em} 2003 \hspace*{2em} 4 \\
**/ \\
create table host ( \\
\hspace*{2em} Host\_ID number, \\
\hspace*{2em} Name text, \\
\hspace*{2em} Nationality text, \\
\hspace*{2em} Age text, \\
\hspace*{2em} primary key (Host\_ID) \\
) \\
/* \\
3 example rows from table host: \\
Host\_ID \hspace*{2em} Name \hspace*{2em} Nationality \hspace*{2em} Age \\
1 \hspace*{2em} Austin Daye \hspace*{2em} United States \hspace*{2em} 43 \\
2 \hspace*{2em} Lloyd Daniels \hspace*{2em} United States \hspace*{2em} 37 \\
3 \hspace*{2em} Kornel David \hspace*{2em} Hungary \hspace*{2em} 31 \\
**/ \\
create table party\_host ( \\
\hspace*{2em} Party\_ID number, \\
\hspace*{2em} Host\_ID number, \\
\hspace*{2em} Is\_Main\_in\_Charge others, \\
\hspace*{2em} primary key (Party\_ID), \\
\hspace*{2em} foreign key (Party\_ID) references party(Party\_ID), \\
\hspace*{2em} foreign key (Host\_ID) references host(Host\_ID) \\
) \\
/* \\
3 example rows from table party\_host: \\
Party\_ID \hspace*{2em} Host\_ID \hspace*{2em} Is\_Main\_in\_Charge \\
1 \hspace*{2em} 1 \hspace*{2em} T \\
8 \hspace*{2em} 7 \hspace*{2em} T \\
6 \hspace*{2em} 10 \hspace*{2em} F \\
**/ \\
Question: What are all the parties? \\
\\
role: assistant \\
content: \\
SELECT * FROM party \\
\\
role: user \\
content: \\
Question: Order them by the number of hosts. \\
\\
role: assistant \\
content: \\
SELECT * FROM party ORDER BY Number\_of\_hosts ASC

\subsection{Self-Defined Edit Rule Style}

role: system \\
content: \\
Given the database schema, you need to translate the question into the SQL query. \\
You can use following operations to edit SQL: \\
1. EditIUE(intersect/union/except, left/right, SQL): Append SQL to the left/right side of the previous SQL with intersect/union/except keyword. Delete the left/right side of the previous SQL with intersect/union/except keyword if SQL is "-". \\
2. EditFromTable(oldTable, newTable): Replace oldTable with newTable in the FROM clause. Add newTable into the FROM clause if oldTable is "-". Delete oldTable from the FROM clause if newTable is "-". \\
3. EditJoinCondition(oldCondition, newCondition): Replace oldCondition with newCondition in the ON clause. Add newCondition into the ON clause if oldCondition is "-". Delete oldCondition from the ON clause if newCondition is "-". \\
4. EditJoinLogicalOperator(and/or): Edit the logical operator in the ON clause. \\
5. EditNestedFromClause(SQL): Edit the nested FROM clause with SQL. Delete the nested FROM clause if SQL is "-". \\
6. EditSelectItem(oldItem, newItem): Replace oldItem with newItem in the SELECT clause. Add newItem into the SELECT clause if oldItem is "-". Delete oldItem from the SELECT clause if newItem is "-". \\
7. EditWhereCondition(oldCondition, newCondition): Replace oldCondition with newCondition in the WHERE clause. Add newCondition into the WHERE clause if oldCondition is "-". Delete oldCondition from the WHERE clause if newCondition is "-". \\
8. EditWhereLogicalOperator(and/or): Edit the logical operator in the WHERE clause. \\
9. EditGroupByColumn(oldColumn, newColumn): Replace oldColumn with newColumn in the GROUP BY clause. Add newColumn into the GROUP BY clause if oldColumn is "-". Delete oldColumn from the GROUP BY clause if newColumn is "-". \\
10. EditHavingCondition(oldCondition, newCondition): Replace oldCondition with newCondition in the HAVING clause. Add newCondition into the HAVING clause if oldCondition is "-". Delete oldCondition from the HAVING clause if newCondition is "-". \\
11. EditHavingLogicalOperator(and/or): Edit the logical operator in the HAVING clause. \\
12. EditOrderByItem(oldItem, newItem): Replace oldItem with newItem in the ORDER BY clause. Add newItem into the ORDER BY clause if oldItem is "-". Delete oldItem from the ORDER BY clause if newItem is "-". \\
13. EditOrder(asc/desc): Edit the order in the ORDER BY clause. \\
14. EditLimit(oldLimit, newLimit): Replace oldLimit with newLimit in the LIMIT clause. Add newLimit into the LIMIT clause if oldLimit is "-". Delete oldLimit from the LIMIT clause if newLimit is "-". \\
\\
role: user \\
content: \\
Database schema: \\
create table party ( \\
\hspace*{2em} Party\_ID number, \\
\hspace*{2em} Party\_Theme text, \\
\hspace*{2em} Location text, \\
\hspace*{2em} First\_year text, \\
\hspace*{2em} Last\_year text, \\
\hspace*{2em} Number\_of\_hosts number, \\
\hspace*{2em} primary key (Party\_ID) \\
) \\
/* \\
3 example rows from table party: \\
Party\_ID \hspace*{2em} Party\_Theme \hspace*{2em} Location \hspace*{2em} First\_year \hspace*{2em} Last\_year \hspace*{2em} Number\_of\_hosts \\
1 \hspace*{2em} Spring \hspace*{2em} Hemkade 48 Zaandam \hspace*{2em} 2000 \hspace*{2em} 2001 \hspace*{2em} 5 \\
2 \hspace*{2em} Club Q-BASE \hspace*{2em} Hemkade 48 Zaandam \hspace*{2em} 2002 \hspace*{2em} 2002 \hspace*{2em} 23 \\
3 \hspace*{2em} Hardcore Resurrection \hspace*{2em} Various locations \hspace*{2em} 2000 \hspace*{2em} 2003 \hspace*{2em} 4 \\
**/ \\
create table host ( \\
\hspace*{2em} Host\_ID number, \\
\hspace*{2em} Name text, \\
\hspace*{2em} Nationality text, \\
\hspace*{2em} Age text, \\
\hspace*{2em} primary key (Host\_ID) \\
) \\
/* \\
3 example rows from table host: \\
Host\_ID \hspace*{2em} Name \hspace*{2em} Nationality \hspace*{2em} Age \\
1 \hspace*{2em} Austin Daye \hspace*{2em} United States \hspace*{2em} 43 \\
2 \hspace*{2em} Lloyd Daniels \hspace*{2em} United States \hspace*{2em} 37 \\
3 \hspace*{2em} Kornel David \hspace*{2em} Hungary \hspace*{2em} 31 \\
**/ \\
create table party\_host ( \\
\hspace*{2em} Party\_ID number, \\
\hspace*{2em} Host\_ID number, \\
\hspace*{2em} Is\_Main\_in\_Charge others, \\
\hspace*{2em} primary key (Party\_ID), \\
\hspace*{2em} foreign key (Party\_ID) references party(Party\_ID), \\
\hspace*{2em} foreign key (Host\_ID) references host(Host\_ID) \\
) \\
/* \\
3 example rows from table party\_host: \\
Party\_ID \hspace*{2em} Host\_ID \hspace*{2em} Is\_Main\_in\_Charge \\
1 \hspace*{2em} 1 \hspace*{2em} T \\
8 \hspace*{2em} 7 \hspace*{2em} T \\
6 \hspace*{2em} 10 \hspace*{2em} F \\
**/ \\
Question 1-1: What are all the parties? \\
\\
role: assistant: \\
content: \\
Let's think step by step. \\
SQL 1-1 can be written directly instead of being edited from previous SQL. \\
So SQL 1-1 is: \\
SELECT * FROM party \\
\\
role: user: \\
content: \\
Question 1-2: Order them by the number of hosts. \\
\\
role: assistant \\
content: \\
Let's think step by step. \\
SQL 1-2 can be edited from SQL 1-1. \\
The previous question asked for a list of all the parties, while the current question asks for the parties to be ordered by the number of hosts they have. \\
Therefore, following edit operations are used: \\
FROM clause: \\
- no change is needed \\
SELECT clause: \\
- no change is needed \\
WHERE clause: \\
- no change is needed \\
GROUP BY clause: \\
- no change is needed \\
ORDER BY clause: \\
- EditOrderByItem(-, party.Number\_of\_hosts) \\
- EditOrder(asc) \\
LIMIT clause: \\
- no change is needed \\
INTERSECT/UNION/EXCEPT: \\
- no change is needed \\
So SQL 1-2 is: \\
SELECT * FROM party ORDER BY Number\_of\_hosts ASC

\subsection{Python Code Style}

role: system \\
content: \\
Given the database schema, you need to translate the question into the SQL query. \\
\\
role: user \\
content: \\
Database schema: \\
create table party ( \\
\hspace*{2em} Party\_ID number, \\
\hspace*{2em} Party\_Theme text, \\
\hspace*{2em} Location text, \\
\hspace*{2em} First\_year text, \\
\hspace*{2em} Last\_year text, \\
\hspace*{2em} Number\_of\_hosts number, \\
\hspace*{2em} primary key (Party\_ID) \\
) \\
/* \\
3 example rows from table party: \\
Party\_ID \hspace*{2em} Party\_Theme \hspace*{2em} Location \hspace*{2em} First\_year \hspace*{2em} Last\_year \hspace*{2em} Number\_of\_hosts \\
1 \hspace*{2em} Spring \hspace*{2em} Hemkade 48 Zaandam \hspace*{2em} 2000 \hspace*{2em} 2001 \hspace*{2em} 5 \\
2 \hspace*{2em} Club Q-BASE \hspace*{2em} Hemkade 48 Zaandam \hspace*{2em} 2002 \hspace*{2em} 2002 \hspace*{2em} 23 \\
3 \hspace*{2em} Hardcore Resurrection \hspace*{2em} Various locations \hspace*{2em} 2000 \hspace*{2em} 2003 \hspace*{2em} 4 \\
**/ \\
create table host ( \\
\hspace*{2em} Host\_ID number, \\
\hspace*{2em} Name text, \\
\hspace*{2em} Nationality text, \\
\hspace*{2em} Age text, \\
\hspace*{2em} primary key (Host\_ID) \\
) \\
/* \\
3 example rows from table host: \\
Host\_ID \hspace*{2em} Name \hspace*{2em} Nationality \hspace*{2em} Age \\
1 \hspace*{2em} Austin Daye \hspace*{2em} United States \hspace*{2em} 43 \\
2 \hspace*{2em} Lloyd Daniels \hspace*{2em} United States \hspace*{2em} 37 \\
3 \hspace*{2em} Kornel David \hspace*{2em} Hungary \hspace*{2em} 31 \\
**/ \\
create table party\_host ( \\
\hspace*{2em} Party\_ID number, \\
\hspace*{2em} Host\_ID number, \\
\hspace*{2em} Is\_Main\_in\_Charge others, \\
\hspace*{2em} primary key (Party\_ID), \\
\hspace*{2em} foreign key (Party\_ID) references party(Party\_ID), \\
\hspace*{2em} foreign key (Host\_ID) references host(Host\_ID) \\
) \\
/* \\
3 example rows from table party\_host: \\
Party\_ID \hspace*{2em} Host\_ID \hspace*{2em} Is\_Main\_in\_Charge \\
1 \hspace*{2em} 1 \hspace*{2em} T \\
8 \hspace*{2em} 7 \hspace*{2em} T \\
6 \hspace*{2em} 10 \hspace*{2em} F \\
**/ \\
Question 1-1: What are all the parties? \\
\\
role: assistant: \\
content: \\
Let's think step by step. \\
SQL 1-1 can be written directly instead of being edited from previous SQL. \\
So SQL 1-1 is: \\
SELECT * FROM party \\
So SQL dict 1-1 is: \\
\{ \\
\hspace*{2em} "from": \{ \\
\hspace*{2em} \hspace*{2em} "tables": [ \\
\hspace*{2em} \hspace*{2em} \hspace*{2em} "party" \\
\hspace*{2em} \hspace*{2em} ] \\
\hspace*{2em} \}, \\
\hspace*{2em} "select": [ \\
\hspace*{2em} \hspace*{2em} "*" \\
\hspace*{2em} ] \\
\} \\
\\
role: user: \\
content: \\
Question 1-2: Order them by the number of hosts. \\
\\
role: assistant \\
content: \\
Let's think step by step. \\
SQL 1-2 can be edited from SQL 1-1. \\
The previous question asked for a list of all the parties, while the current question asks for the parties to be ordered by the number of hosts they have. \\
Therefore, following edit operations are used: \\
sql['order\_by']['columns'].append('party.Number\_of\_hosts') \\
sql['order\_by']['order'] = 'ASC' \\
So SQL 1-2 is: \\
SELECT * FROM party ORDER BY Number\_of\_hosts ASC \\
So SQL dict 1-2 is: \\
\{ \\
\hspace*{2em} "from": \{ \\
\hspace*{2em} \hspace*{2em} "tables": [ \\
\hspace*{2em} \hspace*{2em} \hspace*{2em} "party" \\
\hspace*{2em} \hspace*{2em} ] \\
\hspace*{2em} \}, \\
\hspace*{2em} "select": [ \\
\hspace*{2em} \hspace*{2em} "*" \\
\hspace*{2em} ], \\
\hspace*{2em} "order\_by": \{ \\
\hspace*{2em} \hspace*{2em} "columns": [ \\
\hspace*{2em} \hspace*{2em} \hspace*{2em} "party.Number\_of\_hosts" \\
\hspace*{2em} \hspace*{2em} ], \\
\hspace*{2em} \hspace*{2em} "order": "ASC" \\
\hspace*{2em} \} \\
\}

\subsection{Natural Language Style}

role: system \\
content: \\
Given the database schema, you need to translate the question into the SQL query. \\
\\
role: user \\
content: \\
Database schema: \\
create table party ( \\
\hspace*{2em} Party\_ID number, \\
\hspace*{2em} Party\_Theme text, \\
\hspace*{2em} Location text, \\
\hspace*{2em} First\_year text, \\
\hspace*{2em} Last\_year text, \\
\hspace*{2em} Number\_of\_hosts number, \\
\hspace*{2em} primary key (Party\_ID) \\
) \\
/* \\
3 example rows from table party: \\
Party\_ID \hspace*{2em} Party\_Theme \hspace*{2em} Location \hspace*{2em} First\_year \hspace*{2em} Last\_year \hspace*{2em} Number\_of\_hosts \\
1 \hspace*{2em} Spring \hspace*{2em} Hemkade 48 Zaandam \hspace*{2em} 2000 \hspace*{2em} 2001 \hspace*{2em} 5 \\
2 \hspace*{2em} Club Q-BASE \hspace*{2em} Hemkade 48 Zaandam \hspace*{2em} 2002 \hspace*{2em} 2002 \hspace*{2em} 23 \\
3 \hspace*{2em} Hardcore Resurrection \hspace*{2em} Various locations \hspace*{2em} 2000 \hspace*{2em} 2003 \hspace*{2em} 4 \\
**/ \\
create table host ( \\
\hspace*{2em} Host\_ID number, \\
\hspace*{2em} Name text, \\
\hspace*{2em} Nationality text, \\
\hspace*{2em} Age text, \\
\hspace*{2em} primary key (Host\_ID) \\
) \\
/* \\
3 example rows from table host: \\
Host\_ID \hspace*{2em} Name \hspace*{2em} Nationality \hspace*{2em} Age \\
1 \hspace*{2em} Austin Daye \hspace*{2em} United States \hspace*{2em} 43 \\
2 \hspace*{2em} Lloyd Daniels \hspace*{2em} United States \hspace*{2em} 37 \\
3 \hspace*{2em} Kornel David \hspace*{2em} Hungary \hspace*{2em} 31 \\
**/ \\
create table party\_host ( \\
\hspace*{2em} Party\_ID number, \\
\hspace*{2em} Host\_ID number, \\
\hspace*{2em} Is\_Main\_in\_Charge others, \\
\hspace*{2em} primary key (Party\_ID), \\
\hspace*{2em} foreign key (Party\_ID) references party(Party\_ID), \\
\hspace*{2em} foreign key (Host\_ID) references host(Host\_ID) \\
) \\
/* \\
3 example rows from table party\_host: \\
Party\_ID \hspace*{2em} Host\_ID \hspace*{2em} Is\_Main\_in\_Charge \\
1 \hspace*{2em} 1 \hspace*{2em} T \\
8 \hspace*{2em} 7 \hspace*{2em} T \\
6 \hspace*{2em} 10 \hspace*{2em} F \\
**/ \\
Question 1-1: What are all the parties? \\
\\
role: assistant: \\
content: \\
Let's think step by step. \\
SQL 1-1 can be written directly instead of being edited from previous SQL. \\
So SQL 1-1 is: \\
SELECT * FROM party \\
\\
role: user: \\
content: \\
Question 1-2: Order them by the number of hosts. \\
\\
role: assistant \\
content: \\
Let's think step by step. \\
SQL 1-2 can be edited from SQL 1-1. \\
The previous question asked for a list of all the parties, while the current question asks for the parties to be ordered by the number of hosts they have. \\
Therefore, following edit operations are used: \\
FROM clause: \\
- no change is needed \\
SELECT clause: \\
- no change is needed \\
WHERE clause: \\
- no change is needed \\
GROUP BY clause: \\
- no change is needed \\
ORDER BY clause: \\
- add party.Number\_of\_hosts \\
- change order to ASC \\
LIMIT clause: \\
- no change is needed \\
INTERSECT/UNION/EXCEPT: \\
- no change is needed \\
So SQL 1-2 is: \\
SELECT * FROM party ORDER BY Number\_of\_hosts ASC \\

%% file: acl_latex.bbl
\begin{thebibliography}{29}
\expandafter\ifx\csname natexlab\endcsname\relax\def\natexlab#1{#1}\fi

\bibitem[{Androutsopoulos et~al.(1995)Androutsopoulos, Ritchie, and Thanisch}]{nlidb}
Ion Androutsopoulos, Graeme~D Ritchie, and Peter Thanisch. 1995.
\newblock Natural language interfaces to databases-an introduction.
\newblock \emph{arXiv preprint cmp-lg/9503016}.

\bibitem[{Brown et~al.(2020)Brown, Mann, Ryder, Subbiah, Kaplan, Dhariwal, Neelakantan, Shyam, Sastry, Askell, Agarwal, Herbert-Voss, Krueger, Henighan, Child, Ramesh, Ziegler, Wu, Winter, Hesse, Chen, Sigler, Litwin, Gray, Chess, Clark, Berner, McCandlish, Radford, Sutskever, and Amodei}]{brown2020language}
Tom~B. Brown, Benjamin Mann, Nick Ryder, Melanie Subbiah, Jared Kaplan, Prafulla Dhariwal, Arvind Neelakantan, Pranav Shyam, Girish Sastry, Amanda Askell, Sandhini Agarwal, Ariel Herbert-Voss, Gretchen Krueger, Tom Henighan, Rewon Child, Aditya Ramesh, Daniel~M. Ziegler, Jeffrey Wu, Clemens Winter, Christopher Hesse, Mark Chen, Eric Sigler, Mateusz Litwin, Scott Gray, Benjamin Chess, Jack Clark, Christopher Berner, Sam McCandlish, Alec Radford, Ilya Sutskever, and Dario Amodei. 2020.
\newblock \href {http://arxiv.org/abs/2005.14165} {Language models are few-shot learners}.

\bibitem[{Cai and Wan(2020)}]{cai-wan-2020-igsql}
Yitao Cai and Xiaojun Wan. 2020.
\newblock \href {https://doi.org/10.18653/v1/2020.emnlp-main.560} {{IGSQL}: Database schema interaction graph based neural model for context-dependent text-to-{SQL} generation}.
\newblock In \emph{Proceedings of the 2020 Conference on Empirical Methods in Natural Language Processing (EMNLP)}, pages 6903--6912, Online. Association for Computational Linguistics.

\bibitem[{Cai et~al.(2022)Cai, Li, Hui, Yang, Li, Li, Cao, Li, Huang, Si et~al.}]{star}
Zefeng Cai, Xiangyu Li, Binyuan Hui, Min Yang, Bowen Li, Binhua Li, Zheng Cao, Weijie Li, Fei Huang, Luo Si, et~al. 2022.
\newblock Star: Sql guided pre-training for context-dependent text-to-sql parsing.
\newblock In \emph{Findings of the Association for Computational Linguistics: EMNLP 2022}, pages 1235--1247.

\bibitem[{Chen et~al.(2021{\natexlab{a}})Chen, Tworek, Jun, Yuan, de~Oliveira~Pinto, Kaplan, Edwards, Burda, Joseph, Brockman, Ray, Puri, Krueger, Petrov, Khlaaf, Sastry, Mishkin, Chan, Gray, Ryder, Pavlov, Power, Kaiser, Bavarian, Winter, Tillet, Such, Cummings, Plappert, Chantzis, Barnes, Herbert-Voss, Guss, Nichol, Paino, Tezak, Tang, Babuschkin, Balaji, Jain, Saunders, Hesse, Carr, Leike, Achiam, Misra, Morikawa, Radford, Knight, Brundage, Murati, Mayer, Welinder, McGrew, Amodei, McCandlish, Sutskever, and Zaremba}]{chen2021evaluating}
Mark Chen, Jerry Tworek, Heewoo Jun, Qiming Yuan, Henrique~Ponde de~Oliveira~Pinto, Jared Kaplan, Harri Edwards, Yuri Burda, Nicholas Joseph, Greg Brockman, Alex Ray, Raul Puri, Gretchen Krueger, Michael Petrov, Heidy Khlaaf, Girish Sastry, Pamela Mishkin, Brooke Chan, Scott Gray, Nick Ryder, Mikhail Pavlov, Alethea Power, Lukasz Kaiser, Mohammad Bavarian, Clemens Winter, Philippe Tillet, Felipe~Petroski Such, Dave Cummings, Matthias Plappert, Fotios Chantzis, Elizabeth Barnes, Ariel Herbert-Voss, William~Hebgen Guss, Alex Nichol, Alex Paino, Nikolas Tezak, Jie Tang, Igor Babuschkin, Suchir Balaji, Shantanu Jain, William Saunders, Christopher Hesse, Andrew~N. Carr, Jan Leike, Josh Achiam, Vedant Misra, Evan Morikawa, Alec Radford, Matthew Knight, Miles Brundage, Mira Murati, Katie Mayer, Peter Welinder, Bob McGrew, Dario Amodei, Sam McCandlish, Ilya Sutskever, and Wojciech Zaremba. 2021{\natexlab{a}}.
\newblock \href {http://arxiv.org/abs/2107.03374} {Evaluating large language models trained on code}.

\bibitem[{Chen et~al.(2023{\natexlab{a}})Chen, Lin, Schärli, and Zhou}]{chen2023teaching}
Xinyun Chen, Maxwell Lin, Nathanael Schärli, and Denny Zhou. 2023{\natexlab{a}}.
\newblock \href {http://arxiv.org/abs/2304.05128} {Teaching large language models to self-debug}.

\bibitem[{Chen et~al.(2021{\natexlab{b}})Chen, Chen, Li, Cao, Ma, Wu, and Yu}]{delta}
Zhi Chen, Lu~Chen, Hanqi Li, Ruisheng Cao, Da~Ma, Mengyue Wu, and Kai Yu. 2021{\natexlab{b}}.
\newblock Decoupled dialogue modeling and semantic parsing for multi-turn text-to-sql.
\newblock In \emph{Findings of the Association for Computational Linguistics: ACL-IJCNLP 2021}, pages 3063--3074.

\bibitem[{Chen et~al.(2023{\natexlab{b}})Chen, Chen, White, Mooney, Payani, Srinivasa, Su, and Sun}]{chen-etal-2023-sqledit}
Ziru Chen, Shijie Chen, Michael White, Raymond Mooney, Ali Payani, Jayanth Srinivasa, Yu~Su, and Huan Sun. 2023{\natexlab{b}}.
\newblock \href {https://arxiv.org/abs/2305.13073} {Text-to-sql error correction with language models of code}.
\newblock In \emph{Proceedings of the 61th Annual Meeting of the Association for Computational Linguistics (Volume 2: Short Papers)}, Toronto, Canada. Association for Computational Linguistics.

\bibitem[{Hui et~al.(2021)Hui, Geng, Ren, Li, Li, Sun, Huang, Si, Zhu, and Zhu}]{hui2021dynamic}
Binyuan Hui, Ruiying Geng, Qiyu Ren, Binhua Li, Yongbin Li, Jian Sun, Fei Huang, Luo Si, Pengfei Zhu, and Xiaodan Zhu. 2021.
\newblock \href {http://arxiv.org/abs/2101.01686} {Dynamic hybrid relation network for cross-domain context-dependent semantic parsing}.

\bibitem[{Nan et~al.(2023)Nan, Zhao, Zou, Ri, Tae, Zhang, Cohan, and Radev}]{nan2023enhancing}
Linyong Nan, Yilun Zhao, Weijin Zou, Narutatsu Ri, Jaesung Tae, Ellen Zhang, Arman Cohan, and Dragomir Radev. 2023.
\newblock \href {http://arxiv.org/abs/2305.12586} {Enhancing few-shot text-to-sql capabilities of large language models: A study on prompt design strategies}.

\bibitem[{Ouyang et~al.(2022)Ouyang, Wu, Jiang, Almeida, Wainwright, Mishkin, Zhang, Agarwal, Slama, Ray, Schulman, Hilton, Kelton, Miller, Simens, Askell, Welinder, Christiano, Leike, and Lowe}]{ouyang2022training}
Long Ouyang, Jeff Wu, Xu~Jiang, Diogo Almeida, Carroll~L. Wainwright, Pamela Mishkin, Chong Zhang, Sandhini Agarwal, Katarina Slama, Alex Ray, John Schulman, Jacob Hilton, Fraser Kelton, Luke Miller, Maddie Simens, Amanda Askell, Peter Welinder, Paul Christiano, Jan Leike, and Ryan Lowe. 2022.
\newblock \href {http://arxiv.org/abs/2203.02155} {Training language models to follow instructions with human feedback}.

\bibitem[{Pourreza and Rafiei(2023)}]{pourreza2023dinsql}
Mohammadreza Pourreza and Davood Rafiei. 2023.
\newblock \href {http://arxiv.org/abs/2304.11015} {Din-sql: Decomposed in-context learning of text-to-sql with self-correction}.

\bibitem[{Qi et~al.(2022)Qi, Tang, He, Wan, Cheng, Zhou, Wang, Zhang, and Lin}]{qi-etal-2022-rasat}
Jiexing Qi, Jingyao Tang, Ziwei He, Xiangpeng Wan, Yu~Cheng, Chenghu Zhou, Xinbing Wang, Quanshi Zhang, and Zhouhan Lin. 2022.
\newblock \href {https://doi.org/10.18653/v1/2022.emnlp-main.211} {{RASAT}: Integrating relational structures into pretrained {S}eq2{S}eq model for text-to-{SQL}}.
\newblock In \emph{Proceedings of the 2022 Conference on Empirical Methods in Natural Language Processing}, pages 3215--3229, Abu Dhabi, United Arab Emirates. Association for Computational Linguistics.

\bibitem[{Raffel et~al.(2023)Raffel, Shazeer, Roberts, Lee, Narang, Matena, Zhou, Li, and Liu}]{raffel2023exploring}
Colin Raffel, Noam Shazeer, Adam Roberts, Katherine Lee, Sharan Narang, Michael Matena, Yanqi Zhou, Wei Li, and Peter~J. Liu. 2023.
\newblock \href {http://arxiv.org/abs/1910.10683} {Exploring the limits of transfer learning with a unified text-to-text transformer}.

\bibitem[{Rajkumar et~al.(2022)Rajkumar, Li, and Bahdanau}]{rajkumar2022evaluating}
Nitarshan Rajkumar, Raymond Li, and Dzmitry Bahdanau. 2022.
\newblock \href {http://arxiv.org/abs/2204.00498} {Evaluating the text-to-sql capabilities of large language models}.

\bibitem[{Scholak et~al.(2021)Scholak, Schucher, and Bahdanau}]{scholak-etal-2021-picard}
Torsten Scholak, Nathan Schucher, and Dzmitry Bahdanau. 2021.
\newblock \href {https://doi.org/10.18653/v1/2021.emnlp-main.779} {{PICARD}: Parsing incrementally for constrained auto-regressive decoding from language models}.
\newblock In \emph{Proceedings of the 2021 Conference on Empirical Methods in Natural Language Processing}, pages 9895--9901, Online and Punta Cana, Dominican Republic. Association for Computational Linguistics.

\bibitem[{Vaswani et~al.(2023)Vaswani, Shazeer, Parmar, Uszkoreit, Jones, Gomez, Kaiser, and Polosukhin}]{vaswani2023attention}
Ashish Vaswani, Noam Shazeer, Niki Parmar, Jakob Uszkoreit, Llion Jones, Aidan~N. Gomez, Lukasz Kaiser, and Illia Polosukhin. 2023.
\newblock \href {http://arxiv.org/abs/1706.03762} {Attention is all you need}.

\bibitem[{Wang et~al.(2020)Wang, Ling, Zhou, and Hu}]{wang2020tracking}
Run-Ze Wang, Zhen-Hua Ling, Jing-Bo Zhou, and Yu~Hu. 2020.
\newblock \href {http://arxiv.org/abs/2012.04995} {Tracking interaction states for multi-turn text-to-sql semantic parsing}.

\bibitem[{Wei et~al.(2023)Wei, Wang, Schuurmans, Bosma, Ichter, Xia, Chi, Le, and Zhou}]{wei2023chainofthought}
Jason Wei, Xuezhi Wang, Dale Schuurmans, Maarten Bosma, Brian Ichter, Fei Xia, Ed~Chi, Quoc Le, and Denny Zhou. 2023.
\newblock \href {http://arxiv.org/abs/2201.11903} {Chain-of-thought prompting elicits reasoning in large language models}.

\bibitem[{Xiao et~al.(2022)Xiao, Chai, Zhang, Yan, Li, and Cao}]{cqrsql}
Dongling Xiao, LinZheng Chai, Qian-Wen Zhang, Zhao Yan, Zhoujun Li, and Yunbo Cao. 2022.
\newblock Cqr-sql: Conversational question reformulation enhanced context-dependent text-to-sql parsers.
\newblock In \emph{Findings of the Association for Computational Linguistics: EMNLP 2022}, pages 2055--2068.

\bibitem[{Xu et~al.(2017)Xu, Liu, and Song}]{xu2017sqlnet}
Xiaojun Xu, Chang Liu, and Dawn Song. 2017.
\newblock Sqlnet: Generating structured queries from natural language without reinforcement learning.
\newblock \emph{arXiv preprint arXiv:1711.04436}.

\bibitem[{Yao et~al.(2022)Yao, Zhao, Yu, Shafran, Narasimhan, and Cao}]{react}
Shunyu Yao, Jeffrey Zhao, Dian Yu, Izhak Shafran, Karthik~R Narasimhan, and Yuan Cao. 2022.
\newblock React: Synergizing reasoning and acting in language models.
\newblock In \emph{NeurIPS 2022 Foundation Models for Decision Making Workshop}.

\bibitem[{Yu et~al.(2019{\natexlab{a}})Yu, Zhang, Er, Li, Xue, Pang, Lin, Tan, Shi, Li, Jiang, Yasunaga, Shim, Chen, Fabbri, Li, Chen, Zhang, Dixit, Zhang, Xiong, Socher, Lasecki, and Radev}]{yu-etal-2019-cosql}
Tao Yu, Rui Zhang, Heyang Er, Suyi Li, Eric Xue, Bo~Pang, Xi~Victoria Lin, Yi~Chern Tan, Tianze Shi, Zihan Li, Youxuan Jiang, Michihiro Yasunaga, Sungrok Shim, Tao Chen, Alexander Fabbri, Zifan Li, Luyao Chen, Yuwen Zhang, Shreya Dixit, Vincent Zhang, Caiming Xiong, Richard Socher, Walter Lasecki, and Dragomir Radev. 2019{\natexlab{a}}.
\newblock \href {https://doi.org/10.18653/v1/D19-1204} {{C}o{SQL}: A conversational text-to-{SQL} challenge towards cross-domain natural language interfaces to databases}.
\newblock In \emph{Proceedings of the 2019 Conference on Empirical Methods in Natural Language Processing and the 9th International Joint Conference on Natural Language Processing (EMNLP-IJCNLP)}, pages 1962--1979, Hong Kong, China. Association for Computational Linguistics.

\bibitem[{Yu et~al.(2019{\natexlab{b}})Yu, Zhang, Yasunaga, Tan, Lin, Li, Er, Li, Pang, Chen, Ji, Dixit, Proctor, Shim, Kraft, Zhang, Xiong, Socher, and Radev}]{yu-etal-2019-sparc}
Tao Yu, Rui Zhang, Michihiro Yasunaga, Yi~Chern Tan, Xi~Victoria Lin, Suyi Li, Heyang Er, Irene Li, Bo~Pang, Tao Chen, Emily Ji, Shreya Dixit, David Proctor, Sungrok Shim, Jonathan Kraft, Vincent Zhang, Caiming Xiong, Richard Socher, and Dragomir Radev. 2019{\natexlab{b}}.
\newblock \href {https://doi.org/10.18653/v1/P19-1443} {{SP}ar{C}: Cross-domain semantic parsing in context}.
\newblock In \emph{Proceedings of the 57th Annual Meeting of the Association for Computational Linguistics}, pages 4511--4523, Florence, Italy. Association for Computational Linguistics.

\bibitem[{Zhang et~al.(2023)Zhang, Cao, Chen, Xu, and Yu}]{zhang2023actsql}
Hanchong Zhang, Ruisheng Cao, Lu~Chen, Hongshen Xu, and Kai Yu. 2023.
\newblock \href {http://arxiv.org/abs/2310.17342} {Act-sql: In-context learning for text-to-sql with automatically-generated chain-of-thought}.

\bibitem[{Zhang et~al.(2019)Zhang, Yu, Er, Shim, Xue, Lin, Shi, Xiong, Socher, and Radev}]{zhang-etal-2019-editing}
Rui Zhang, Tao Yu, Heyang Er, Sungrok Shim, Eric Xue, Xi~Victoria Lin, Tianze Shi, Caiming Xiong, Richard Socher, and Dragomir Radev. 2019.
\newblock \href {https://doi.org/10.18653/v1/D19-1537} {Editing-based {SQL} query generation for cross-domain context-dependent questions}.
\newblock In \emph{Proceedings of the 2019 Conference on Empirical Methods in Natural Language Processing and the 9th International Joint Conference on Natural Language Processing (EMNLP-IJCNLP)}, pages 5338--5349, Hong Kong, China. Association for Computational Linguistics.

\bibitem[{Zheng et~al.(2022)Zheng, Wang, Dong, Wang, and Li}]{zheng-etal-2022-hie}
Yanzhao Zheng, Haibin Wang, Baohua Dong, Xingjun Wang, and Changshan Li. 2022.
\newblock \href {https://doi.org/10.18653/v1/2022.findings-acl.236} {{HIE}-{SQL}: History information enhanced network for context-dependent text-to-{SQL} semantic parsing}.
\newblock In \emph{Findings of the Association for Computational Linguistics: ACL 2022}, pages 2997--3007, Dublin, Ireland. Association for Computational Linguistics.

\bibitem[{Zhong et~al.(2020)Zhong, Lewis, Wang, and Zettlemoyer}]{zhong-etal-2020-grounded}
Victor Zhong, Mike Lewis, Sida~I. Wang, and Luke Zettlemoyer. 2020.
\newblock \href {https://doi.org/10.18653/v1/2020.emnlp-main.558} {Grounded adaptation for zero-shot executable semantic parsing}.
\newblock In \emph{Proceedings of the 2020 Conference on Empirical Methods in Natural Language Processing (EMNLP)}, pages 6869--6882, Online. Association for Computational Linguistics.

\bibitem[{Zhong et~al.(2017)Zhong, Xiong, and Socher}]{zhongSeq2SQL2017}
Victor Zhong, Caiming Xiong, and Richard Socher. 2017.
\newblock Seq2sql: Generating structured queries from natural language using reinforcement learning.
\newblock \emph{CoRR}, abs/1709.00103.

\end{thebibliography}
